\documentclass[11pt]{article}

\usepackage[final]{acl}
\usepackage{times}
\usepackage{latexsym}

\usepackage[T1]{fontenc}
\usepackage[utf8]{inputenc}

\usepackage{microtype}

\usepackage{inconsolata}

\usepackage{graphicx}

\usepackage{algorithm}
\usepackage{algorithmic}
\usepackage{enumitem}
\usepackage{amsmath}
\usepackage{titleref}
\usepackage{multirow}
\usepackage[most]{tcolorbox}
\title{From Texts to Scores: Tracing the Emergence of Essay Quality Representations in Large Language Models}

\author{
 \textbf{Jiaxu Zuo\textsuperscript{1}},~~
 \textbf{Mu You\textsuperscript{2}},~~
 \textbf{Kaixin Lan\textsuperscript{1}},~~
 \textbf{Tao Fang\textsuperscript{2}},~~
 \textbf{Yujia Huo\textsuperscript{3}},~~
 \textbf{Henghua Shen\textsuperscript{2}},
 \\
 \textbf{Lidia S. Chao\textsuperscript{1}},~~
 \textbf{Derek F. Wong \textsuperscript{1}}\thanks{~Corresponding Author}
\\
 \textsuperscript{1}NLP$^2$CT Lab, Department of Computer and Information Science, University of Macau
 \\
 \texttt{\{mc45440,lidiasc,derekfw\}@um.edu.mo, nlp2ct.kaixin@gmail.com}
 \\
 \textsuperscript{2}Institute of International Language Services Studies, Macau Millennium College
 \\
\texttt{youmuafonso@gmail.com, taofang@mmc.edu.mo, henghua.shen@dal.ca}
\\
 \textsuperscript{3}School of Data Science and Information Engineering, Guizhou Minzu University
 \\
 \texttt{huo.yujia@gzmu.edu.cn}
}

\begin{document}
\maketitle
% \begin{abstract}
% With the rapid advancement of Large Language Models (LLMs), the research paradigm in Automated Essay Scoring (AES) is undergoing a profound transformation. Current research is no longer limited to merely enhancing scoring performance but has expanded to providing diagnostic feedback. However, the black-box nature of LLMs raises concerns about their reliability and hinders their application in educational practice. To interpret their internal scoring mechanisms, a critical question must be addressed: Does the essay scoring capability of LLMs stem from merely memorizing superficial correlations between text and scores, or from a genuine understanding of essay quality? To address this, we systematically analyze the hidden states of six models across two English datasets (ASAP++, CSEE) and one Portuguese dataset (ENEM). Our findings provide evidence for the latter: LLMs learn linear representations of essay quality across multiple scales. These representations exhibit robustness to variations in LLM prompts and consistency across different essay prompts. Furthermore, we identify individual “essay scoring neurons” that reliably encode essay quality. Finally, we reveal that the distribution of these neurons within the network is associated with essay length.
% \end{abstract}

\begin{abstract}
Recent advances in Large Language Models (LLMs) have substantially transformed Automated Essay Scoring (AES), yet the internal mechanisms underlying LLM-based scoring remain poorly understood. In this work, we systematically analyze the hidden representations of eight LLMs across two English essay datasets (ASAP++, CSEE) and one Portuguese dataset (ENEM). Using linear probing, cross-prompt generalization, dimensionality reduction, and neuron-level analyses, we find consistent evidence that essay quality information is encoded in a linearly accessible form within LLM representations. These representations emerge progressively across layers, remain robust across prompting strategies, and partially transfer across essay prompts despite differences in scoring rubrics. In addition, nonlinear probes provide only marginal and inconsistent improvements over linear probes, suggesting that most essay quality information is already linearly decodable. We further identify individual ``essay scoring neurons'' whose activations strongly correlate with essay scores and whose behavior is sensitive to targeted intervention. Moreover, the layer-wise distribution of these neurons systematically shifts with essay length, with longer essays relying more heavily on deeper layers. Overall, our findings provide evidence that LLMs encode structured representations related to essay quality and offer new insights into the interpretability of LLM-based AES systems.
\end{abstract}

\section{Introduction}

Automated Essay Scoring (AES) aims to provide scalable and consistent evaluation of student writing. Traditional AES methods have largely followed two paradigms. Prompt-specific models are trained and evaluated on essays from the same essay prompt\footnote{To avoid confusion between essay prompts (writing tasks assigned to students) and LLM prompts (inputs to large language models), the unqualified term \textit{prompt} in this paper refers to essay prompts.}, achieving strong in-domain performance but often generalizing poorly to unseen prompts \cite{rudner2002automated,miltsakaki2004evaluation,yannakoudakis2011new,rodriguez2019languagemodelsautomatedessay,nadeem-etal-2019-automated,yang-etal-2020-enhancing,uto-etal-2020-neural,alikaniotis-etal-2016-automatic,dong-etal-2017-attention,taghipour-ng-2016-neural}. Cross-prompt approaches improve transferability through domain adaptation and generalization techniques \cite{ridley2020prompt,li-ng-2024-conundrums,chen-li-2023-pmaes,wang2025making,zhang2025pairwise}, but they still rely heavily on annotated data and typically underperform compared with prompt-specific systems.

Recent advances in Large Language Models (LLMs) have substantially changed this landscape. With carefully designed prompts, LLMs can perform essay scoring in zero-shot or few-shot settings \cite{mizumoto2023exploring,yancey2023rating,escalante2023ai,stahl2024exploring}, reducing the dependence on labeled datasets. Moreover, unlike conventional AES systems that mainly output numerical scores, LLMs can also provide diagnostic feedback and personalized comments, enabling richer forms of writing assessment. However, despite these advantages, LLM-based AES still faces important challenges. Their scoring performance often remains unstable compared with strong supervised AES systems \cite{lee-etal-2024-unleashing}. In addition, as LLMs remain inherently black-box systems, their internal decision-making processes are opaque, and their outputs are highly sensitive to LLM prompt design \cite{han2024llm}. These limitations raise concerns about reliability and trustworthiness in educational applications, particularly in high-stakes assessment settings.

A central open question is therefore how LLMs internally represent essay quality. In particular, it remains unclear whether LLMs derive their scoring ability primarily from superficial statistical cues or whether they learn structured representations that capture higher-level aspects of writing quality. Understanding this distinction is important not only for interpretability, but also for evaluating the robustness and generalizability of LLM-based AES systems.

In this work, we investigate the internal representations underlying LLM-based AES through representation- and neuron-level analyses. We analyze eight models on two English essay datasets and one Portuguese dataset. Through linear probing, cross-prompt generalization, dimensionality reduction, and neuron intervention experiments, we study how essay quality information is represented across model layers and neurons.

Our results show that essay quality information is progressively constructed across layers and is largely linearly decodable from hidden representations. These representations remain relatively stable across prompting strategies and partially transfer across essay prompts despite differences in scoring rubrics. We further identify individual ``essay-scoring neurons'' that strongly correlate with essay scores and exhibit sensitivity to targeted intervention. Finally, we find that the layer-wise distribution of these neurons systematically shifts with essay length, suggesting that longer essays rely more heavily on deeper-layer computations. Together, these findings provide new insights into the internal mechanisms underlying LLM-based AES and contribute toward more interpretable and trustworthy intelligent scoring systems.

\section{Related Work}
\label{sec:related_Work}
\subsection{Automated Essay Scoring}
Early prompt-specific AES models, based on handcrafted features or neural networks, required labeled data for each new essay prompt \cite{miltsakaki2004evaluation, yannakoudakis2011new, alikaniotis-etal-2016-automatic, dong-etal-2017-attention, rodriguez2019languagemodelsautomatedessay}. To improve generalization, cross-prompt methods were later proposed \cite{ridley2020prompt,chen-li-2023-pmaes,li-ng-2024-conundrums,wang2025making,zhang2025pairwise}. More recently, LLM-based zero-shot AES has emerged, enabling essay scoring without labeled data \cite{mizumoto2023exploring, yancey2023rating,escalante2023ai,stahl2024exploring}. Early approaches relied on simple rubric-based prompting, while later methods such as Multi-Trait Specification \cite{lee-etal-2024-unleashing} introduced fine-grained, trait-level evaluation. However, direct scoring remains sensitive to LLM prompt design and prone to bias. RRecent work by \citet{shibata2025lces} addresses these issues by reformulating AES as a pairwise essay comparison task, improving robustness at the cost of greater computational overhead and reliance on unlabeled data. 

% Despite these advances, the internal mechanisms through which LLMs derive essay scores remain largely unexplored.

% Early prompt-specific models, reliant on handcrafted features or neural networks, require costly labeled data for each new essay topic \cite{rudner2002automated, miltsakaki2004evaluation, yannakoudakis2011new, rodriguez2019languagemodelsautomatedessay, nadeem-etal-2019-automated, yang-etal-2020-enhancing, uto-etal-2020-neural, alikaniotis-etal-2016-automatic, dong-etal-2017-attention, taghipour-ng-2016-neural}. To improve generalization, cross-prompt methods were developed to train models that transfer across different writing prompts \cite{ridley2020prompt,chen-li-2023-pmaes,li-ng-2024-conundrums,wang2025making,zhang2025pairwise}. Most recently, LLM-based zero-shot AES has emerged, which generates scores without any task-specific training essays \cite{mizumoto2023exploring, yancey2023rating,escalante2023ai,stahl2024exploring}. Initial approaches used simple rubric-based prompting, followed by more structured frameworks like Multi-Trait Specification \cite{lee-etal-2024-unleashing} that perform fine-grained, trait-level evaluation. However, these direct scoring methods can exhibit bias and sensitivity to instructions. The latest advancements, \citet{shibata2025lces} address this by reframing scoring as a comparative task to reduce bias, albeit with increased computational cost, representing an ongoing effort to balance accuracy, generalizability, and reliability in automated assessment.

\subsection{Interpretability and Probing}
A major direction in interpretability research concerns identifying what information is encoded in model representations and how that information supports downstream tasks. Probing methods have become one of the dominant approaches for this purpose. In probing, external classifiers are trained on hidden representations to predict linguistic, semantic, or task-related attributes, under the assumption that successful prediction indicates that the relevant information is encoded in the model \cite{ettinger-etal-2016-probing, belinkov-glass-2019-analysis}. Probing studies have been used to analyze a wide range of properties, including syntax, morphology, factual knowledge, and reasoning abilities across pretrained language models. However, subsequent work has questioned whether probe performance alone provides reliable evidence about representation quality. In particular, expressive probes may recover task signals independently of the structure of the underlying representation, making it difficult to distinguish information genuinely encoded by the model from information introduced by the probe itself \cite{hewitt-liang-2019-designing,belinkov-2022-probing}. These limitations have motivated a broader shift toward studying the structure, geometry, and dynamics of representations rather than relying exclusively on probing accuracy.

Recent interpretability research therefore increasingly focuses on understanding how representations are organized internally and how they support model computation. Prior work has examined geometric properties of contextual embeddings \cite{rogers2020primerbertologyknowbert}, investigated the emergence of linear features and feature superposition in deep networks \cite{elhage2022toymodelssuperposition}, and developed mechanistic interpretability techniques aimed at identifying neurons, attention heads, or circuits associated with particular behaviors \cite{olah2020zoom, räuker2023transparentaisurveyinterpreting}. Collectively, these studies move beyond the question of whether information exists in a representation toward understanding how information is distributed, transformed, and utilized during inference. 
Despite these advances, interpretability research in LLM-based AES remains limited. Existing work has largely concentrated on prompting strategies that generate explanations or formative feedback for users \cite{xiao2025human}, while comparatively little attention has been paid to the internal representations underlying essay evaluation itself. While concurrent work has demonstrated that LLM activations can serve as effective features for cross-prompt scoring \cite{chi2026activations}, it remains unclear how these models structurally encode and utilize essay-quality signals during inference.

\begin{figure*}[t]
    \centering
\includegraphics[width=1\linewidth]{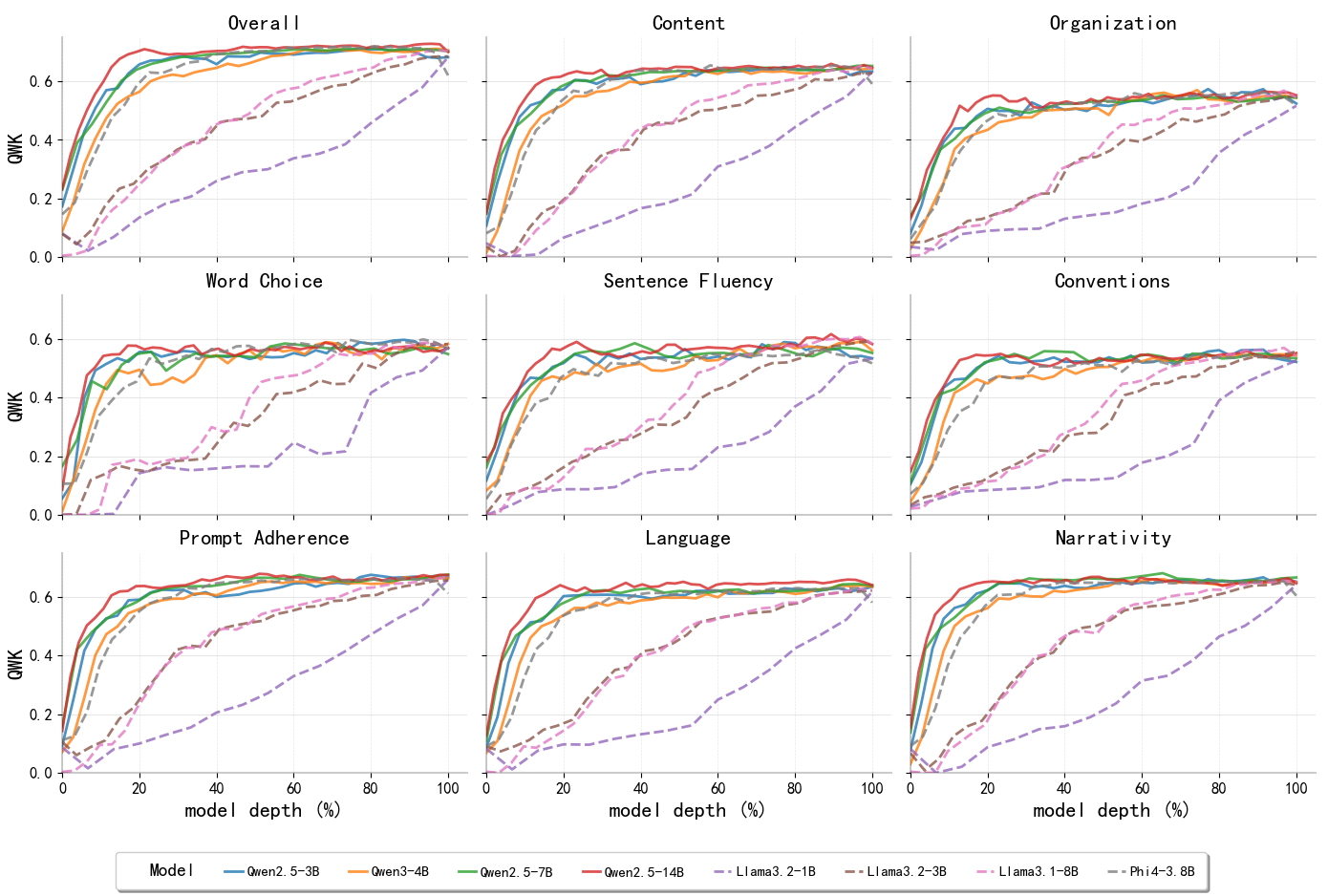}
\caption{Average QWK scores of linear probes across all essay prompts in ASAP++. Each subplot corresponds to an essay trait and shows probe performance across layers for different models.}
\label{fig:linear}
\end{figure*}
\section{Approach}
% Given a dataset of $n$ essays $E = \{e_1, e_2, \dots, e_n\}$ and their corresponding human-annotated target scores $Y = \{y_1, y_2, \dots, y_n\}$ (which can be overall scores or specific trait scores such as content and organization), we aim to investigate whether the essay quality information is encoded within the LLM's internal representations.

% For each dataset, we input all essays into the model and extract the hidden state activations (residual stream) corresponding to the last token of each essay across all layers. This process yields an activation matrix $A \in R^{n \times d_{model}}$ for each layer.

% For each corpus, we input all essays into the model and extract the hidden state activations (residual stream) corresponding to the last token of each essay across all layers. For a corpus of $n$ essays, where $\mathbf{h}^{(l)}_{i, L_i}$ denotes the hidden state activation at layer $l$ for the last token of the $i$-th essay, this process yields an activation matrix for each layer, formulated as:
% \begin{equation}
% \mathbf{A}^{(l)} = \begin{bmatrix} \mathbf{h}^{(l)}_{1, L_1} \\ \mathbf{h}^{(l)}_{2, L_2} \\ \vdots \\ \mathbf{h}^{(l)}_{n, L_n} \end{bmatrix} \in \mathbb{R}^{n \times d_{\text{model}}}
% \end{equation}

Given a dataset of $n$ essays $E = \{e_1, e_2, \dots, e_n\}$ and their corresponding human-annotated target scores $Y = \{y_1, y_2, \dots, y_n\}$ (which can be overall scores or trait scores), we feed all essays into the model and extract the hidden state activations (i.e., residual stream representations) corresponding to the final token of each essay across all layers. Let $H_i^{(l)} \in R^{L_i \times d_{\text{model}}}$ denote the hidden state matrix for the $i$-th essay at layer $l$, where $L_i$ is the sequence length. The hidden state activation corresponding to the last token is extracted as:
\[h_{i, L_i}^{(l)} = H_i^{(l)}[-1, :] \in R^{1 \times d_{\text{model}}}\]

Collecting these representations across all essays yields the activation matrix for layer $l$:
\[A^{(l)} = \begin{bmatrix} h_{1, L_1}^{(l)} \\ h_{2, L_2}^{(l)} \\ \vdots \\ h_{n, L_n}^{(l)} \end{bmatrix} \in R^{n \times d_{\text{model}}}\]

To examine whether LLM representations encode essay quality information, we adopt standard probing methodologies \cite{alain2018understanding,belinkov2022probing}, which aim to assess whether target labels associated with annotated inputs can be recovered from model representations using simple supervised predictors. Specifically, given an activation matrix $A^{(l)}$ and target scores \( Y \), we train a linear ridge regression probe defined as

\[
\hat{W}
=
\arg\min_{W}
\left\|Y - A^{(l)}W\right\|_2^2
+
\lambda \left\|W\right\|_2^2,
\]
where \(\lambda\) denotes the regularization coefficient. The closed-form solution is given by
\[
\hat{W}
=
\left(A^{(l)\top}A^{(l)} + \lambda I\right)^{-1}
A^{(l)\top}Y.
\]

Using the learned probe parameters, predictions are obtained as
\[
\hat{Y} = A^{(l)}\hat{W}.
\]

Strong generalization performance on out-of-sample data suggests that essay quality information is linearly decodable from the underlying model representations. However, consistent with prior work \citep{ravichander2021probing,gurnee2023language}, successful probing does not necessarily imply that the base model itself utilizes these representations during inference. In all experiments, the regularization parameter \(\lambda\) is selected via efficient leave-one-out cross-validation performed on the probe training set \citep{hastie2009elements}.

\section{Experiments}
\subsection{LLMs}
We evaluate eight instruction-tuned LLMs from the Llama-3.1/3.2 \cite{grattafiori2024llama}, Qwen2.5/3 \cite{qwen2.5,qwen3technicalreport}, and Phi-4 \cite{microsoft2025phi4minitechnicalreportcompact} families: Llama-3.2-1B-Instruct, Llama-3.2-3B-Instruct, Llama-3.1-8B-Instruct, Qwen2.5-3B-Instruct, Qwen3-4B-Instruct-2507, Qwen2.5-7B-Instruct, Qwen2.5-14B-Instruct, and Phi-4-mini-instruct. Ranging from 1B to 14B parameters, these models represent diverse architectures and training strategies, supporting the generalizability of our findings.

% \begin{figure*}[h]
%     \centering
%     \includegraphics[width=1\linewidth]{LLM_prompt.png}
%     \caption{QWK scores of linear probes trained on the overall score of the ASAP dataset across different LLM prompts and essay prompts on Llama-3.1-8B-Instruct.}
%     \label{fig:LLM_prompt}
% \end{figure*}

\subsection{Datasets and Evaluation Metrics}
We conduct experiments on two English essay-scoring datasets, ASAP++ \cite{mathias2018asap++} and CSEE\footnote{\url{https://catalog.ldc.upenn.edu/LDC2014T06}} \cite{xiao2025human}, as well as a Portuguese dataset ENEM\footnote{\url{https://github.com/kamel-usp/aes_enem}} \cite{silveira2024new}. We include CSEE to mitigate potential data leakage concerns, as the dataset was released in 2025, after the training cutoff dates of the evaluated models. Detailed descriptions of all datasets are provided in Appendix~\ref{sec:dataset}.
% A description of these datasets is provided below and the statistics of these datasets are in Appendix~\ref{sec:dataset}.  

% \paragraph{ASAP++} is an extension of the ASAP\footnote{\url{https://www.kaggle.com/c/asap-aes/data}} dataset which comprises 12,978 essays written by students in grades 7-10. These essays are produced in response to eight different prompts, which vary in genre and scoring criteria. Each essay has an overall score and 8 trait scores.

% \paragraph{CSEE} is carefully curated in collaboration with 29 high schools in China, encompassing a total of 13,372 student essays responding to two distinct prompts used in final exams. Each essay has an overall score and 3 trait scores. 

% \paragraph{ENEM} is divided into two subsets: \textbf{Source A}, with 386 essays including full supporting texts validated by experts, serves as a high-quality benchmark; \textbf{Source B}, with 3,200 essays, is mainly used for model pretraining and augmentation \cite{silveira-etal-2024-new}. Each essay has an overall score and 5 trait scores. We only use source A for experiments.

Following prior AES research \cite{dong-etal-2017-attention,li-ng-2024-conundrums}, we evaluate model performance using the Quadratic Weighted Kappa (QWK) metric \cite{cohen1960coefficient}. Consistent with standard practice in prompt-specific AES settings \cite{dong-etal-2017-attention,xiao2025human}, we split each dataset into 80\% training data and 20\% testing data.

All experimental details (including model training and probe configurations) are provided in Appendix~\ref{sec:setting}.

\begin{figure*}[t]
    \centering
\includegraphics[width=1\linewidth]{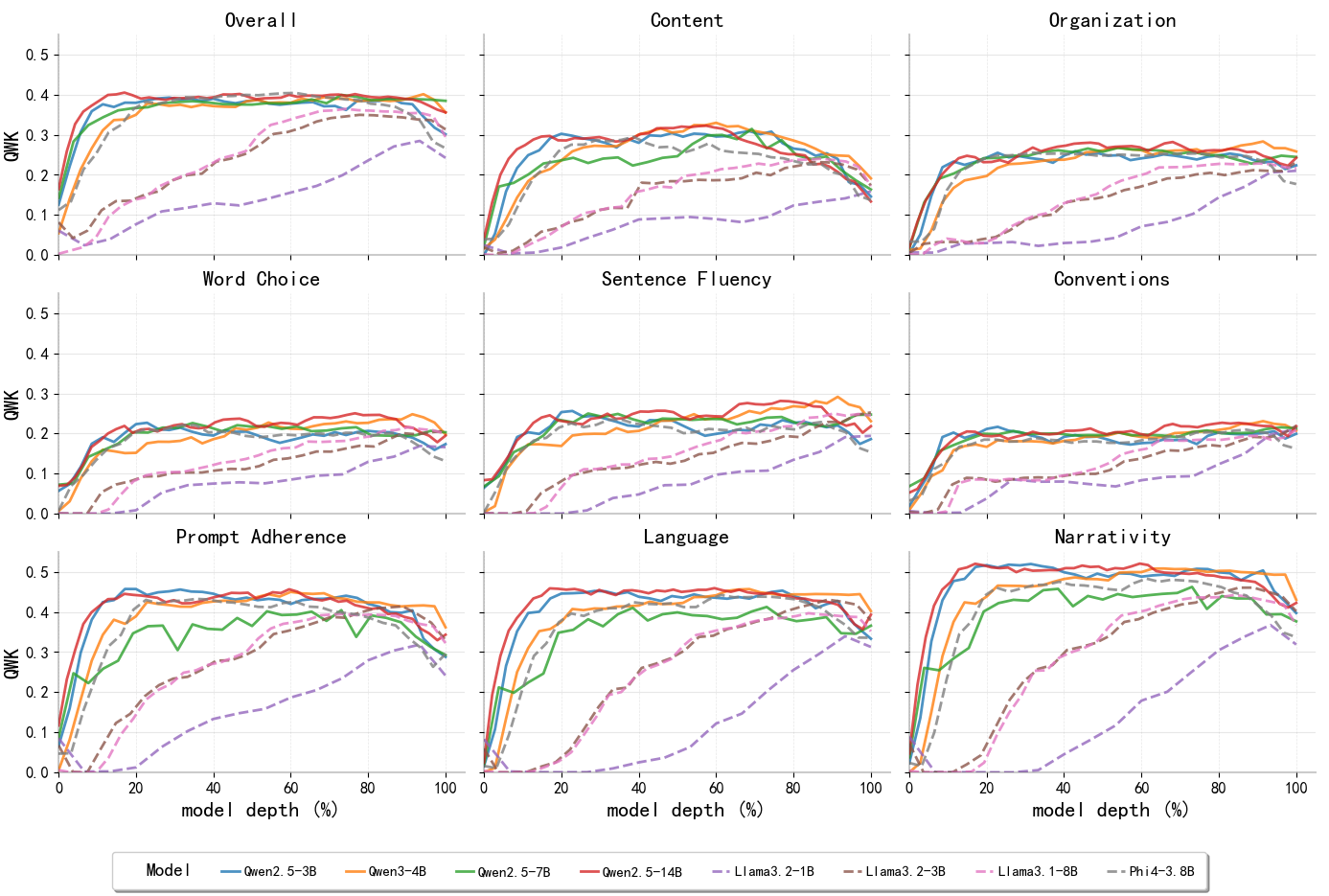}
    % \vspace{-20pt} % 减少图片下方的空白
\caption{Average QWK scores of linear probes on ASAP++ under cross-prompt settings. Each subplot corresponds to an essay trait and shows probe performance across layers for different models.}    \label{fig:cross_linear}
\end{figure*}

\subsection{Results}
\textbf{Essay Quality in Representations.} As shown in Figure~\ref{fig:linear}, linear probes exhibit similar trends across models and traits. Essay quality information becomes increasingly accessible in deeper layers, while final-layer performance remains broadly comparable across different models. Larger models tend to encode essay quality information more rapidly in earlier layers, leading to steeper initial performance gains, but their final peak performance differs only marginally from that of smaller models.

The results also reveal two interesting patterns. First, Llama models exhibit behavior that differs markedly from that of Qwen and Phi models. Although their final-layer QWK scores (hence representation quality) are comparable, Qwen and Phi models reach saturation at around 20\% of model depth and subsequently plateau, whereas Llama models continue improving steadily all the way until the final layers. At present, we can only offer a tentative hypothesis for this phenomenon: it may stem from differences in training data quality, as prior work has shown that high-quality training data can substantially shape model capabilities, enabling smaller models to rival larger ones \cite{gunasekar2023textbooksneed,zhang2024tinyllamaopensourcesmalllanguage}.
Second, the probes consistently predict overall essay scores more accurately than individual trait scores. This suggests that the models capture coarse-grained essay quality representations more effectively than fine-grained trait-specific ones.

\noindent\textbf{Linear Decodability.} We compare linear ridge regression probes with more expressive nonlinear MLP probes of the form $W_2\mathrm{ReLU}(W_1x + b_1) + b_2$, using 256 hidden neurons (see Appendix~\ref{sec:Appendix C}). Across traits, nonlinear probes provide only marginal and inconsistent improvements in QWK over linear probes. This suggests that essay quality information is largely linearly decodable from the hidden representations, i.e., a linear readout is sufficient to recover most of the task-relevant signal. This finding is consistent with prior work in interpretability research supporting the linear representation hypothesis, which proposes that features in neural networks can be recovered by projecting activations onto corresponding feature directions \cite{mikolov2013linguistic,olah2020zoom,elhage2022toymodelssuperposition,gurnee2023language}.
% \paragraph{Linear representations.} As shown in Table~\ref{table:linear}, we compare the performance of linear ridge regression probes with that of more expressive nonlinear MLP probes (with the structure $W2ReLU(W1x+b1)+b2$, containing 256 neurons). For any traits, the improvement in the QWK metric using nonlinear probes is minimal. This aligns with findings in interpretability research, where increasing evidence supports the linear representation hypothesis---that features in neural networks are represented linearly, meaning the presence or strength of a feature can be read by projecting relevant activations onto a feature vector \cite{mikolov2013linguistic,olah2020zoom,elhage2022toymodelssuperposition,gurnee2023language}. This suggests that essay quality features are represented in a linear manner.

\begin{figure*}[t]
    \centering
    % \vspace{-100pt} % 减少图片上方的空白
    \includegraphics[width=1\linewidth]{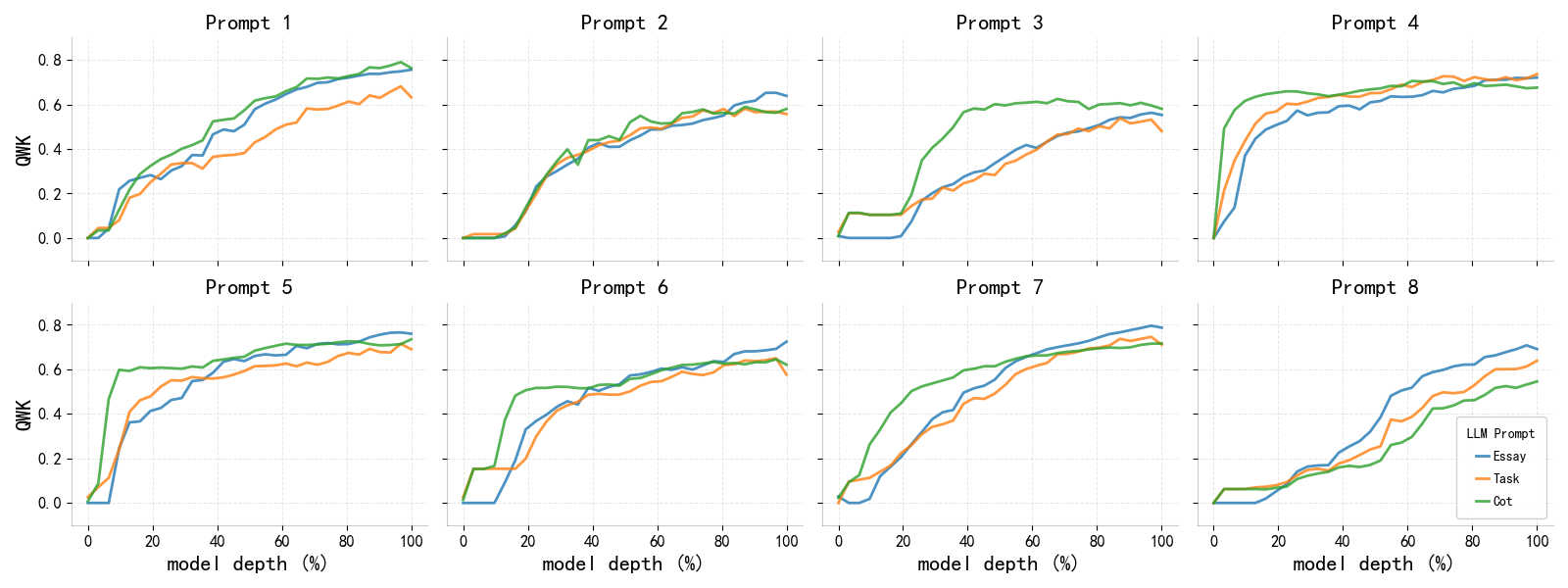}
    % \vspace{-90pt} % 减少图片下方的空白
\caption{QWK scores of linear probes trained on overall essay scores in ASAP++ using Llama-3.1-8B-Instruct. Each subplot corresponds to an essay prompt and shows probe performance across three prompting strategies.}    \label{fig:LLM_prompt}
\end{figure*}

\noindent\textbf{LLM Prompt Robustness.} We analyze the sensitivity of essay quality representations to LLM prompt design. In practice, models are typically provided with both the essay and task instructions, and prior work has shown that prompting strategies can affect scoring performance \cite{stahl2024exploring, lee-etal-2024-unleashing}. We consider three prompt variants: Essay (essay only), Task (instructions + essay), and Chain-of-Thought (CoT) (instructions eliciting step-by-step reasoning + essay) (see Appendix~\ref{sec:llm_prompt}). 
We evaluate these strategies using Llama-8B on ASAP++ for overall score prediction. As shown in Figure~\ref{fig:LLM_prompt}, prompt variations lead to only marginal differences in essay quality representations. In most cases, CoT yields faster convergence and earlier saturation of probe performance, suggesting that explicit reasoning instructions may better elicit the model’s latent scoring knowledge. However, for Prompt 8, which contains longer essays, CoT performs worst, followed by Task. This may be due to the additional instructional text introducing noise for longer inputs, which interferes with the formation of stable essay quality representations.

\section{Discussion and Analysis}

\subsection{Cross-Prompt Generalization}
The previous section demonstrated that both overall and trait-specific essay scores can be linearly reconstructed from internal activations of later LLM layers. However, this result alone does not imply that the model explicitly represents essay quality in the directions identified by the probe, as the probe may instead exploit linear combinations of more primitive features already present in the representations \cite{gurnee2023language}.

To evaluate cross-prompt generalization, we retrain the linear probe on the ASAP++ dataset using the same cross-prompt partitioning strategy as in prior work \cite{ridley2020prompt, li-ng-2024-conundrums}. Specifically, for each target prompt, all remaining prompts are used as training data, and evaluation is performed on the held-out prompt.

As shown in Figure~\ref{fig:cross_linear}, the resulting performance trends closely match those in Figure~\ref{fig:linear}. The observed performance drop relative to in-prompt training is likely attributable to differences in scoring rubrics and prompt-specific distribution shifts. Importantly, probe performance remains substantially above chance (QWK = 0), indicating that a non-trivial portion of the essay quality signal is shared and linearly accessible across diverse prompts, despite prompt-dependent variation in how it is encoded.

\begin{figure*}[h]
    \centering
\includegraphics[width=1\linewidth]{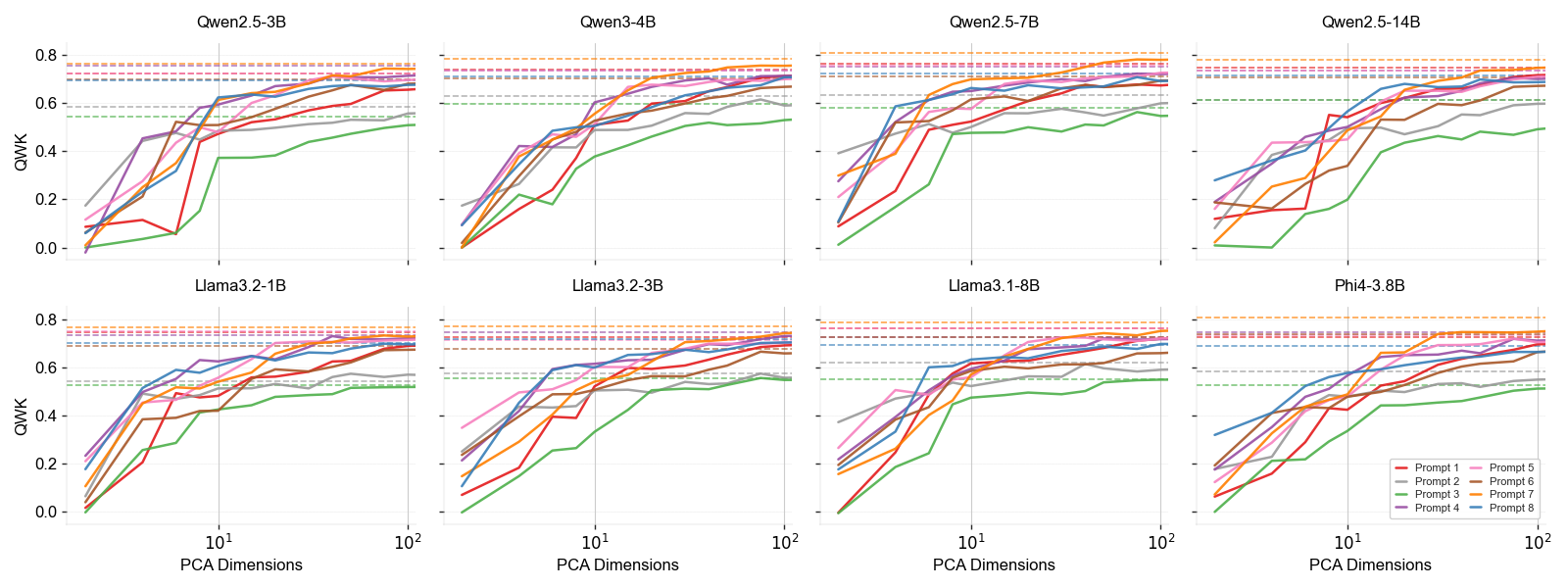}
\caption{QWK scores across PCA dimensionality settings for each model. Dotted lines denote probes trained on full-dimensional activations.}
\label{fig:pca}
\end{figure*}

% \begin{figure*}[h]
%     \centering
%         \vspace{-250pt} % 减少图片下方的空白
% \includegraphics[width=1\linewidth]{pca.pdf}
%     \vspace{-230pt} % 减少图片下方的空白
%     \caption{QWK scores of probes trained on activations projected onto the $k$ largest principal components, compared to training on full activations, for each essay prompt and model on the overall score of the ASAP dataset.}
%     \label{fig:pca}
% \end{figure*}

\subsection{Dimensionality Reduction}
Although the probes we employ are linear, they operate in the full hidden dimensionality $d_{\text{model}}$ (ranging from 2048 to 5120 for models with 1B to 14B parameters), which still allows for non-trivial capacity and potential memorization. As an additional robustness check, we use Principal Component Analysis (PCA) \cite{shlens2014tutorialprincipalcomponentanalysis} to project the activation space onto its top $k$ principal components and train linear probes in this reduced subspace, thereby reducing the number of parameters by 2–3 orders of magnitude.

Figure \ref{fig:pca} reports performance of probes trained to predict overall essay scores on the ASAP++ dataset across varying values of $k$, and compares them with full-dimensional probes. Results for Spearman correlation (see Appendix \ref{sec:pca_sp}) show that these coefficients increase more rapidly with $k$ than QWK. This difference is expected, as Spearman correlation depends only on the rank ordering of predictions, whereas QWK additionally penalizes deviations in absolute score calibration. Overall, these results suggest that low-dimensional projections already capture substantial rank-relevant information about essay quality, while higher-dimensional components appear more important for improving calibration of absolute score predictions.

\begin{figure*}[h]
    \centering
\includegraphics[width=1\linewidth]{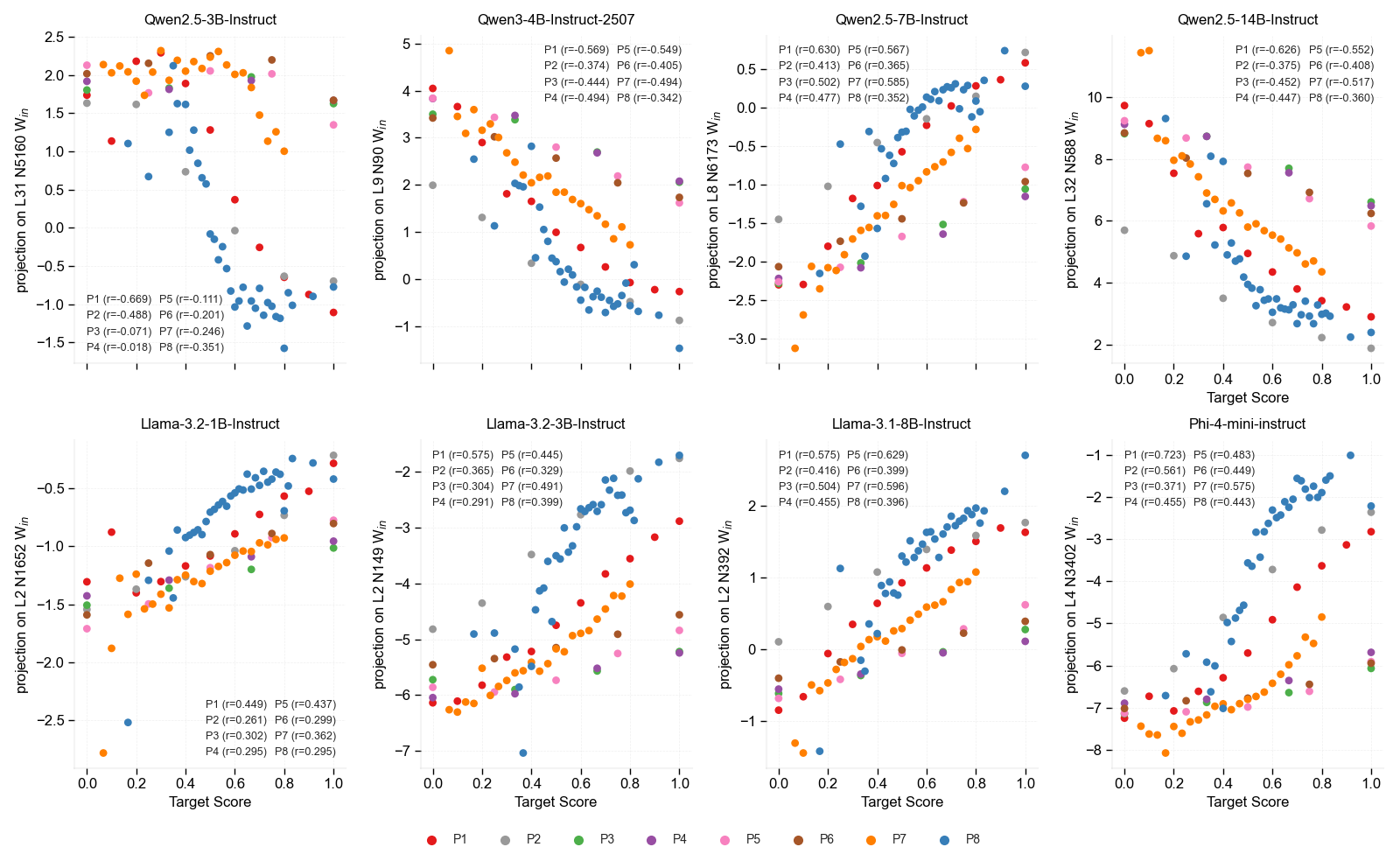}
\caption{Essay scoring neurons in each model. Spearman correlations between neuron-weight projections and true essay scores are shown for each ASAP++ prompt. Each point denotes the average projection value for a target score.}    \label{fig:neuron}
\end{figure*}

\subsection{Essay Scoring Neurons}

While the previous experimental results are informative, they provide only indirect evidence and do not establish whether the LLMs explicitly utilize the feature directions identified by the probes. To address this more directly, we identify individual neurons whose input or output weight vectors exhibit high cosine similarity with the probe-derived feature directions. Specifically, we focus on the overall score prediction for Prompt 1 in the ASAP++ dataset and compute the Spearman correlation between ground-truth scores and neuron activation values.

As shown in Figure~\ref{fig:neuron}, projecting the activation data onto the weights of these most similar neurons reveals that certain individual neurons are themselves highly sensitive to overall essay scores. In other words, some neurons can serve as effective standalone feature probes. Notably, neurons identified as most relevant for Prompt 1 also transfer to other prompts, maintaining substantial Spearman correlations, which suggests a degree of cross-prompt consistency in these representations.

If feature directions learned by supervised linear probes approximate the upper bound of the model’s linearly decodable essay-related information, then the performance of individual neurons can be viewed as a lower bound. It is important to note that such features are generally expected to be distributed across multiple neurons in a superpositioned manner, making single-neuron analysis inherently limited \cite{elhage2022toymodelssuperposition}. Nevertheless, the existence of individual neurons, learned solely via the next-token prediction objective, that align with essay-scoring behavior provides evidence that the model encodes and utilizes features related to essay quality. We further conduct neuron intervention experiments (see Appendix~\ref{sec:Intervention}), which suggest that these neurons are more sensitive to intervention and play a more prominent role in the model’s essay scoring behavior than typical neurons.

\begin{figure*}[h]
    \centering
\includegraphics[width=1\linewidth]{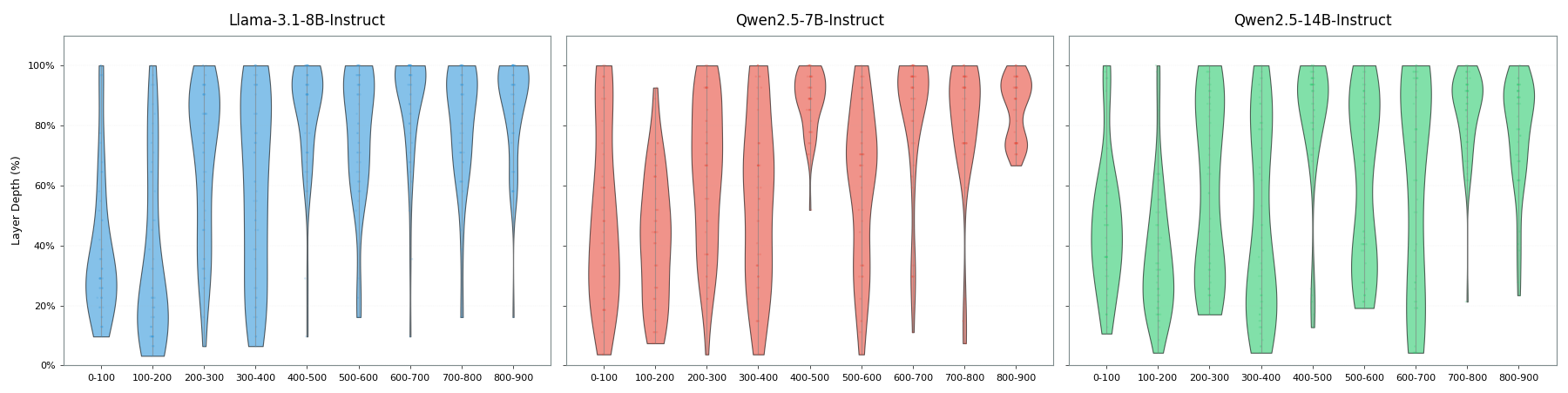}
    \caption{Distribution of the top 50 essay scoring neurons for the overall score of Prompt 8 in the ASAP++ dataset across different essay length intervals and models.}
    \label{fig:layer_length}
\end{figure*}

\subsection{Neuron Distribution}
% Following the identification of feature neurons critical to essay scoring, we further analyzed their distribution patterns across network layers to investigate differences in the model’s internal mechanisms when processing different essay prompts. All models exhibit a consistent pattern in the distribution of these neurons (see the Appendix~\ref{sec:Neuron} for details). Specifically, for essays with shorter length (prompt 3, 4, 5, 6), the neurons are distributed in the early to middle layers. In contrast, for essays with greater length (prompt 1, 2, 7, 8), the neurons are distributed in the middle to later layers. Furthermore, compared to smaller models, the neuron distribution in larger models tends to be positioned in earlier layers.
Following the identification of neurons critical to essay scoring, we further investigate how internal processing varies across essay prompts by analyzing their distribution across network layers. On the ASAP++ dataset using Llama-3.1-8B-Instruct, Qwen2.5-7B-Instruct, and Qwen2.5-14B-Instruct, we observe a consistent distribution pattern for these neurons (see Appendix~\ref{sec:Neuron}). Specifically, for shorter essays (Prompts 3--6), the neurons are predominantly located in earlier to middle layers, whereas for longer essays (Prompts 1, 2, 7, and 8), they are more frequently found in middle to later layers. In addition, larger models tend to exhibit earlier emergence of these neurons compared to smaller models.

To further examine the point at which this distribution shifts with respect to text length, we conduct a finer-grained analysis on Prompt 8. We partition essays into 100-word bins to study how neuron distribution varies across different length ranges. To ensure sufficient samples per bin, we additionally generate essays using LLMs (see Appendix~\ref{sec:llm_essay}). As shown in Figure~\ref{fig:layer_length}, the distribution of essay scoring neurons begins to shift toward later layers once text length reaches approximately 200 words, consistently across all three models.

Interestingly, this shift appears to align with cognitive accounts of reading under increased load. For humans, longer texts introduce extended syntactic dependencies, increasing working memory demands and requiring deeper integration of information \cite{gibson1998linguistic}. Similarly, in neural networks, earlier layers tend to capture local and syntactic patterns, while deeper layers are more involved in long-range and discourse-level integration \cite{tenney2019bertrediscoversclassicalnlp}. From this perspective, the observed shift toward deeper layers for longer essays suggests that the model may adaptively recruit higher-layer computations to accommodate increased integration demands associated with increased essay length.

\section{Conclusion}
In this work, we investigate how large language models internally represent essay quality for automated essay scoring. Through extensive probing experiments, we show that both overall and trait-specific essay scores can be effectively decoded from hidden representations using simple linear probes, with essay quality information becoming increasingly accessible in deeper layers. While larger models tend to encode such information earlier in the network, final-layer performance remains broadly comparable across model families. Furthermore, nonlinear probes provide only marginal improvements over linear ones, suggesting that essay quality information is largely linearly decodable from LLM representations.

We further demonstrate that these representations are robust across prompting strategies and partially transferable across essay prompts, despite differences in scoring rubrics and prompt-specific distribution shifts. Dimensionality reduction experiments additionally show that low-dimensional subspaces already preserve substantial rank-relevant information about essay quality, indicating that these signals are not solely dependent on high-capacity probe parameterization.

Beyond representation-level analysis, we identify individual neurons whose activations strongly correlate with essay scores and whose weight vectors align with probe-derived feature directions. Neuron intervention experiments further suggest that these neurons play a more prominent role in the model's essay scoring behavior than typical neurons. Moreover, we observe systematic shifts in the layer-wise distribution of essay scoring neurons as essay length increases, with longer essays relying more heavily on deeper layers. This pattern suggests that LLMs may recruit deeper computations to accommodate the increased integration demands associated with longer text inputs.

Overall, our findings provide evidence that LLMs encode structured and linearly accessible representations related to essay quality, extending beyond superficial statistical cues. More broadly, this work contributes toward bridging black-box performance and mechanistic interpretability in AES. Future work may explore how these representations and neurons can be leveraged to improve scoring robustness, controllability, and interpretability in educational applications.

% Our investigation conclusively demonstrates that the essay scoring capability of LLMs stems from learning linear representations that encode substantive essay quality features, rather than merely memorizing superficial linguistic statistics. These representations demonstrate robustness across different LLM prompt settings and exhibit consistency across different essay prompts. We further identified specific essay scoring neurons that are highly sensitive to signals of these  representations, and through neuron intervention experiments, confirm the importance of these neurons in modeling essay. Moreover, the study reveals that the distribution of these key neurons within the network is linked to essay length, providing new insights into the process of internal representation formation.

% These findings bridge the gap between black-box performance and explainability in AES, laying a foundation for developing the next generation of high-performance, high-trust, and explainable intelligent scoring systems. Future work could explore how to leverage these linear representations and key neurons to improve model scoring performance, enhance interpretability, and design more reliable evaluation frameworks.

\section*{Limitations}
While this study provides insights into the internal mechanisms of LLM-based AES, several limitations remain.

\begin{itemize}
    \item[(i)] \textbf{Limited Model Scale:} Our experiments focused on open-source LLMs, ranging from 1B to 14B parameters, and did not include large-scale commercial models such as GPT-4 or Llama-3.1-70B-Instruct. Since many capabilities emerge with scale, it remains unclear whether our findings generalize to substantially larger models.

    \item[(ii)] \textbf{Limited Language Coverage:} Experiments were conducted on two English datasets (ASAP++ and CSEE) and one Portuguese dataset (ENEM). Although the results suggest some degree of cross-lingual transferability, evaluating only one non-English language limits the generalizability of our conclusions. Essay scoring criteria may vary across cultural contexts, writing conventions, and educational systems, requiring broader multilingual evaluation.

    \item[(iii)] \textbf{Limited Mechanistic Analysis:} Although we identified individual ``essay scoring neurons'', feature superposition may limit the interpretability of single-neuron analysis. In addition, our intervention experiments were restricted to individual neurons and therefore do not capture potential interactions among multiple neurons or higher-level scoring circuits.
\end{itemize}

\section*{Acknowledgements}
This work was supported in part by the Science and Technology Development Fund of Macau SAR (Grant Nos. FDCT/0007/2024/AKP, EF2024-00185-FST), the UM and UMDF (Grant Nos. MYRG-GRG2024-00165-FST-UMDF, MYRG-GRG2025-00236-FST), the Tencent AI Lab Rhino-Bird Research Program (Grant No. EF2023-00151-FST), the Dr. Stanley Ho Medical Development Foundation (Grant No. SHMDF-AI/2026/001), and the National Natural Science Foundation of China (Grant No. 62266013). This work was performed in part at SICC which is supported by SKL-IOTSC, and HPCC supported by ICTO of the University of Macau.

\bibliography{custom}

\appendix
\section*{Appendix}
\section{Datasets}
\label{sec:dataset}
\paragraph{ASAP++} is an extension of the ASAP\footnote{\url{https://www.kaggle.com/c/asap-aes/data}} dataset which comprises 12,978 essays written by students in grades 7-10. These essays are produced in response to eight different prompts, which vary in genre and scoring criteria. Each essay has an overall score and 8 trait scores. The descriptive statistics of ASAP++ are outlined in Table \ref{tab:asap_stats}.

\paragraph{CSEE} is carefully curated in collaboration with 29 high schools in China, encompassing a total of 13,372 student essays responding to two distinct prompts used in final exams. Each essay has an overall score and 3 trait scores. The evaluation of these essays was carried out by highly experienced English teachers following the scoring guidelines of the Chinese National College Entrance Examination. Scoring was comprehensively assessed across three critical dimensions: Content, Language, and Structure, with an Overall Score ranging from 0 to 20. The descriptive statistics of CSEE are outlined in Table \ref{tab:csee_stats}.

\paragraph{ENEM} comprises argumentative essays written by Brazilian students in response to a variety of socially relevant prompts. Collected from public online platforms simulating the Brazilian National High School Exam (ENEM), these essays are annotated following the official ENEM scoring rubric. The rubric evaluates five aspects (C1–C5, C1: fluency, C2: writing style, C3: argumentation quality, C4: proper use of textual connectors, and C5: quality of the solution to the prompt’s problem), each scored on a scale from 0 to 200 in increments of 20, resulting in a total score out of 1000. 

The dataset is divided into two subsets: \textbf{Source A}, with 386 essays including full supporting texts validated by experts, serves as a high-quality benchmark; \textbf{Source B}, with 3,200 essays, is mainly used for model pretraining and augmentation \cite{silveira-etal-2024-new}. We only use source A for experiments.

% \paragraph{CSEE} is carefully curated in collaboration with 29 high schools in China, encompassing a total of 13,372 student essays responding to two distinct prompts used in final exams. Each essay has an overall score and 3 trait scores. 

% \paragraph{ENEM} is divided into two subsets: \textbf{Source A}, with 386 essays including full supporting texts validated by experts, serves as a high-quality benchmark; \textbf{Source B}, with 3,200 essays, is mainly used for model pretraining and augmentation \cite{silveira-etal-2024-new}. Each essay has an overall score and 5 trait scores (C1-C5, C1: ). We only use source A for experiments.

\begin{table*}[b]
\centering
\resizebox{1\textwidth}{!}{%
\begin{tabular}{ccccccc}
\hline
\multirow{2}{*}{Prompt ID} & \multirow{2}{*}{No. of Essays} & \multirow{2}{*}{Avg. Len.} & \multirow{2}{*}{Genre} & \multirow{2}{*}{Attributes} & \multicolumn{2}{c}{Score Range} \\
& & & & &  Overall & Attribute \\
\hline
1 & 1,783 & 418 & ARG & Cont, Org, WC, SF, Conv & 2 - 12 & 1 - 6 \\
2 & 1,800 & 427 & ARG & Cont, Org, WC, SF, Conv & 0 - 6 & 1 - 6 \\
3 & 1,726 & 123 & RES & Cont, PA, Lan, Nar & 0 - 3 & 0 - 3 \\
4 & 1,772 & 105 & RES & Cont, PA, Lan, Nar & 0 - 3 & 0 - 3 \\
5 & 1,805 & 140 & RES & Cont, PA, Lan, Nar & 0 - 4 & 0 - 4 \\
6 & 1,800 & 172 & RES & Cont, PA, Lan, Nar & 0 - 4 & 0 - 4 \\
7 & 1,569 & 199 & NAR & Cont, Org, Conv & 0 - 30 & 0 - 6 \\
8 & 723 & 701 & NAR & Cont, Org, WC, SF, Conv & 0 - 60 & 2 - 12 \\
\hline
\end{tabular}
}
\caption{Statistics of ASAP++. Abbreviations: Cont (Content), Org (Organization), WC (Word Choice), SF (Sentence Fluency), Conv (Conventions), PA (Prompt Adherence), Lan (Language), Nar (Narrativity). `Avg. Len.' refers to the average essay length in tokens, calculated using the NLTK toolkit (\url{https://www.nltk.org/}).}
\label{tab:asap_stats}
\end{table*}

\begin{table}[h]
\centering
\begin{tabular}{cc}
\hline
\multicolumn{2}{c}{Statistics of CSEE} \\
\hline
\# of schools & 29 \\
\# of essay prompts & 2 \\
\# of student essays & 13,372 \\
avg. essay length & 124.74 \\
avg. Overall score & 10.72 \\
avg. Content score & 4.13 \\
avg. Language score & 4.05 \\
avg. Structure score & 2.55 \\
\hline
\end{tabular}
\caption{Descriptive statistics of Chinese Student English Essay (CSEE) dataset \cite{xiao2025human}.}
\label{tab:csee_stats}
\end{table}

\section{Experiments Settings}
\label{sec:setting}
To ensure reproducibility, all experiments are conducted with a fixed random seed of $42$. For the linear probe, we use ridge regression with built-in cross-validation (\texttt{RidgeCV}), searching the regularization strength $\alpha$ over 12 logarithmically spaced values in the range $[10^3, 10^{4.5}]$, while retaining the cross-validation scores. 

For the nonlinear probe, we adopt a single-hidden-layer multi-layer perceptron with a hidden dimension of $256$. The multi-layer perceptron is trained using AdamW with mean squared error loss, a fixed learning rate of $1\times10^{-3}$, and a batch size of $4096$. We tune weight decay over $\{0.01, 0.03, 0.1, 0.3\}$. Training is run for up to $200$ epochs with early stopping based on a $10\%$ held-out validation split from the training set; training is stopped if the validation loss does not improve for $10$ consecutive epochs.

\section{Linear vs. Nonlinear Probes}
\label{sec:Appendix C}
In Table~\ref{table:linear}, we present the average QWK scores of linear and nonlinear probes on ASAP++ essay traits, averaged across all prompts and models at full ($100\%$) layer depth.

\begin{table*}[t] 
\centering
\resizebox{1\textwidth}{!}{%
\begin{tabular}{cccccccccccc}
\hline
\textbf{Model} & \textbf{Probe} & \textbf{Overall} & \textbf{Cont} & \textbf{Org} & \textbf{WC} & \textbf{SF} & \textbf{Conv} & \textbf{PA} & \textbf{Lan} & \textbf{Nar} & \textbf{Avg.} \\ \hline
\multirow{2}{*}{Llama3.2-1B} & Linear & 0.682 & 0.632 & 0.516 & 0.571 & 0.534 & 0.527 & 0.659 & 0.617 & 0.641 & 0.598 \\ 
& Nonlinear & 0.691 & 0.637 & 0.570 & 0.541 & 0.544 & 0.540 & 0.663 & 0.639 & 0.657 & 0.609 \\ 
\hline
\multirow{2}{*}{Llama3.2-3B} & Linear & 0.685 & 0.627 & 0.549 & 0.583 & 0.593 & 0.558 & 0.657 & 0.622 & 0.641 & 0.613 \\ 
& Nonlinear & 0.653 & 0.620 & 0.544 & 0.536 & 0.576 & 0.555 & 0.631 & 0.616 & 0.634 & 0.596 \\ 
\hline
\multirow{2}{*}{Llama3.1-8B} & Linear & 0.704 & 0.644 & 0.545 & 0.572 & 0.585 & 0.541 & 0.668 & 0.621 & 0.651 & 0.615 \\ 
& Nonlinear & 0.667 & 0.631 & 0.529 & 0.520 & 0.564 & 0.539 & 0.586 & 0.588 & 0.646 & 0.585 \\ 
\hline
\multirow{2}{*}{Phi4-3.8B} & Linear & 0.620 & 0.591 & 0.518 & 0.565 & 0.517 & 0.528 & 0.612 & 0.581 & 0.604 & 0.571 \\ 
& Nonlinear & 0.644 & 0.617 & 0.552 & 0.569 & 0.468 & 0.496 & 0.613 & 0.565 & 0.574 & 0.567 \\ 
\hline
\multirow{2}{*}{Qwen2.5-3B} & Linear & 0.682 & 0.633 & 0.524 & 0.568 & 0.534 & 0.522 & 0.666 & 0.635 & 0.652 & 0.602 \\ 
& Nonlinear & 0.684 & 0.623 & 0.520 & 0.562 & 0.527 & 0.527 & 0.666 & 0.626 & 0.660 & 0.599 \\ 
\hline
\multirow{2}{*}{Qwen3-4B} & Linear & 0.705 & 0.644 & 0.544 & 0.583 & 0.561 & 0.544 & 0.663 & 0.635 & 0.650 & 0.614 \\ 
& Nonlinear & 0.692 & 0.643 & 0.579 & 0.542 & 0.602 & 0.594 & 0.653 & 0.636 & 0.635 & 0.620 \\ 
\hline
\multirow{2}{*}{Qwen2.5-7B} & Linear & 0.703 & 0.652 & 0.544 & 0.549 & 0.553 & 0.534 & 0.676 & 0.639 & 0.666 & 0.613 \\ 
& Nonlinear & 0.691 & 0.638 & 0.543 & 0.527 & 0.562 & 0.544 & 0.682 & 0.606 & 0.659 & 0.606 \\ 
\hline
\multirow{2}{*}{Qwen2.5-14B} & Linear & 0.700 & 0.646 & 0.552 & 0.570 & 0.583 & 0.552 & 0.674 & 0.640 & 0.646 & 0.618 \\ 
& Nonlinear & 0.643 & 0.602 & 0.521 & 0.594 & 0.590 & 0.566 & 0.652 & 0.613 & 0.615 & 0.599 \\ 
\hline
\end{tabular}
}
\caption{Average QWK scores of linear and nonlinear probes on ASAP++ essay traits, averaged across all prompts and models at full (100\%) layer depth. Abbreviations: Cont (Content), Org (Organization), WC (Word Choice), SF (Sentence Fluency), Conv (Conventions), PA (Prompt Adherence), Lan (Language), Nar (Narrativity).}
\label{table:linear}
\end{table*}

\section{LLM Prompting Templates}
\label{sec:llm_prompt}
We present two examples based on the prompt template, corresponding to the Task prompt and the CoT prompt, respectively.

\subsection{Task Prompt}
\begin{tcolorbox}[
    enhanced,
    colback=white,
    colframe=black,
    boxrule=0.8pt,
    arc=3pt,
    width=0.47\textwidth, % 稍微缩短整体宽度
    breakable,
]
As an English teacher, your primary responsibility is to evaluate the writing quality of essays written by middle school students, with evaluation measured on a scale from \{min\_score\} to \{max\_score\}.\\\relax
[Essay]\\
\{essay\}\\
(end of [Essay])
\end{tcolorbox}

\subsection{CoT Prompt}
\begin{tcolorbox}[
    enhanced,
    colback=white,
    colframe=black,
    boxrule=0.8pt,
    arc=3pt,
    width=0.47\textwidth, % 稍微缩短整体宽度
    breakable,
]
As an English teacher, your primary responsibility is to evaluate the writing quality of essays written by middle school students. During the assessment process, you will be provided with an essay. First, you should provide comprehensive and concrete feedback that is closely linked to the content of the essay. It is essential to avoid offering generic remarks that could be applied to any piece of writing. To create a compelling evaluation for both the student and fellow experts, you should reference specific content of the essay to substantiate your assessment. Next, your evaluation should culminate in assigning an overall score to the student's essay, measured on a scale from \{min\_score\} to \{max\_score\}, where higher score should reflect a higher level of writing quality. It's crucial to tailor your evaluation criteria to be well-suited for middle school level writing, taking into account the developmental stage and capabilities of these students.\\\relax
[Essay]\\
\{essay\}\\
(end of [Essay])
\end{tcolorbox}

\begin{figure*}[b]
    \centering
    \includegraphics[width=1\linewidth]{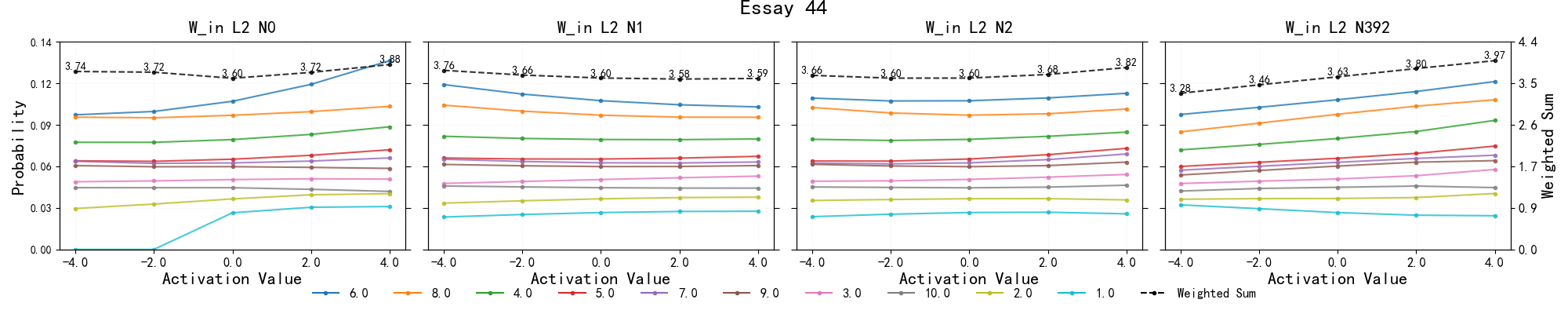}
    \hfill
    \includegraphics[width=1\linewidth]{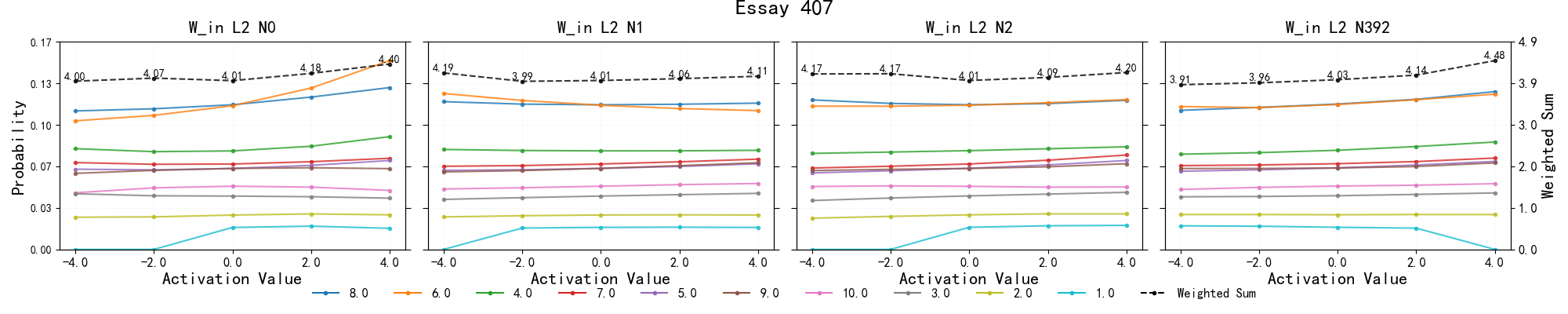}
    \hfill
    \includegraphics[width=1\linewidth]{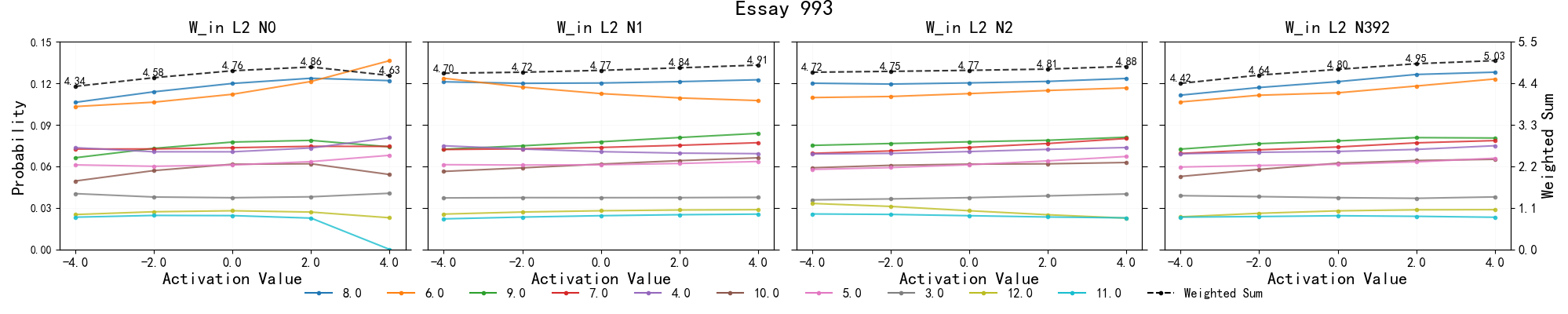}
    \hfill
    \includegraphics[width=1\linewidth]{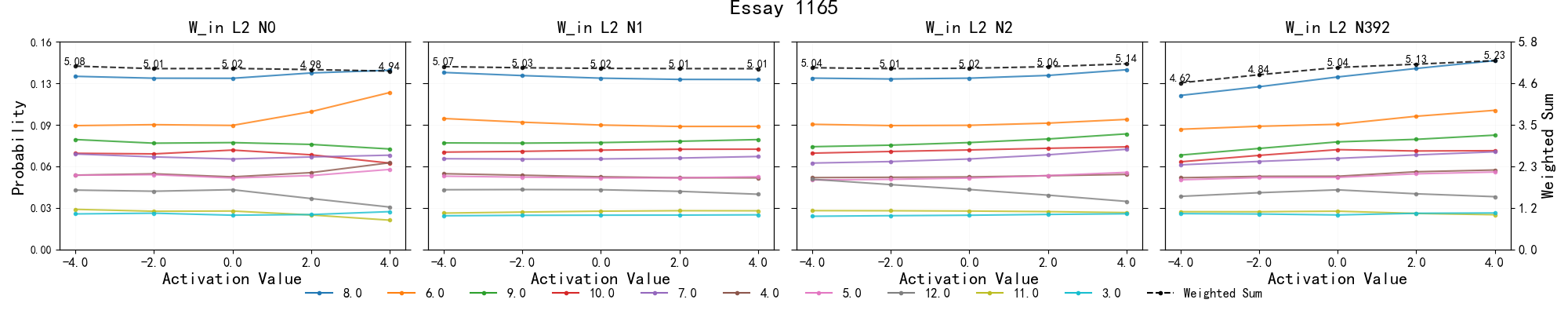}
    \hfill
    \includegraphics[width=1\linewidth]{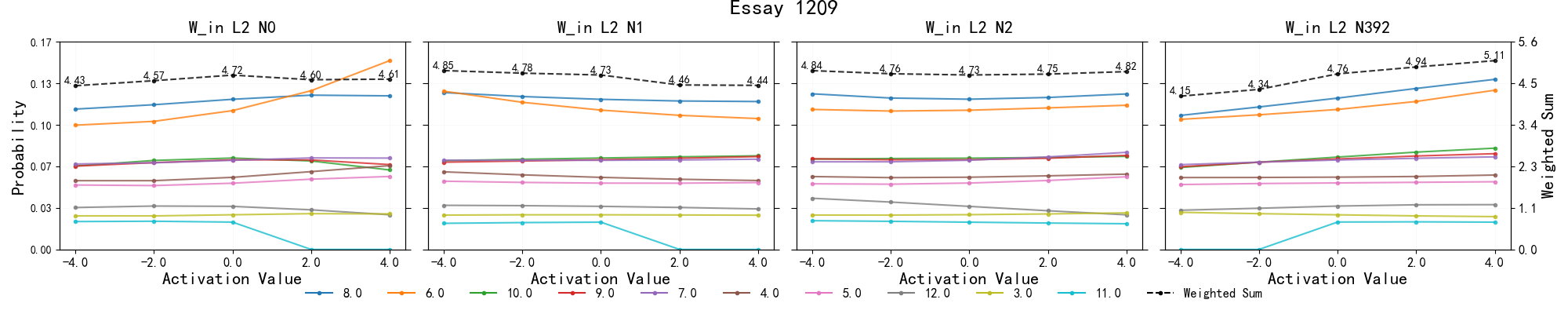}
    \caption{When the essay scoring neuron ($L2.N392.W_{in}$) is fixed to specific values, the prediction results for five different essays from prompt 1 of the ASAP dataset are compared with the prediction results from three random neurons in the same layer (L2.[0-2]) of the Llama-3.1-8B-Instruct model. We also calculate the weighted sum of top 10 tokens when the essay scoring neuron is fixed to different specific values.}
    \label{fig:Intervention}
\end{figure*}

\begin{figure*}[h]
    \centering
\includegraphics[width=1\linewidth]{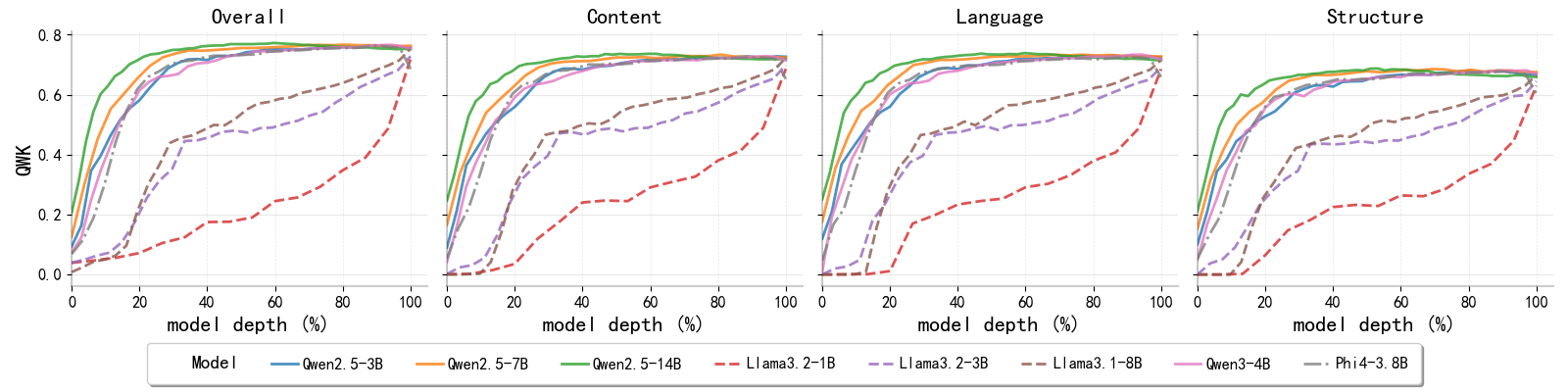}
    \caption{Average QWK scores of linear probes trained on CSEE across all essay prompts. Each subplot corresponds to a essay trait and shows probe performance across layers for different models.}
    \label{fig:CSEE}
\end{figure*}

\begin{figure*}[t]
    \centering
\includegraphics[width=1\linewidth]{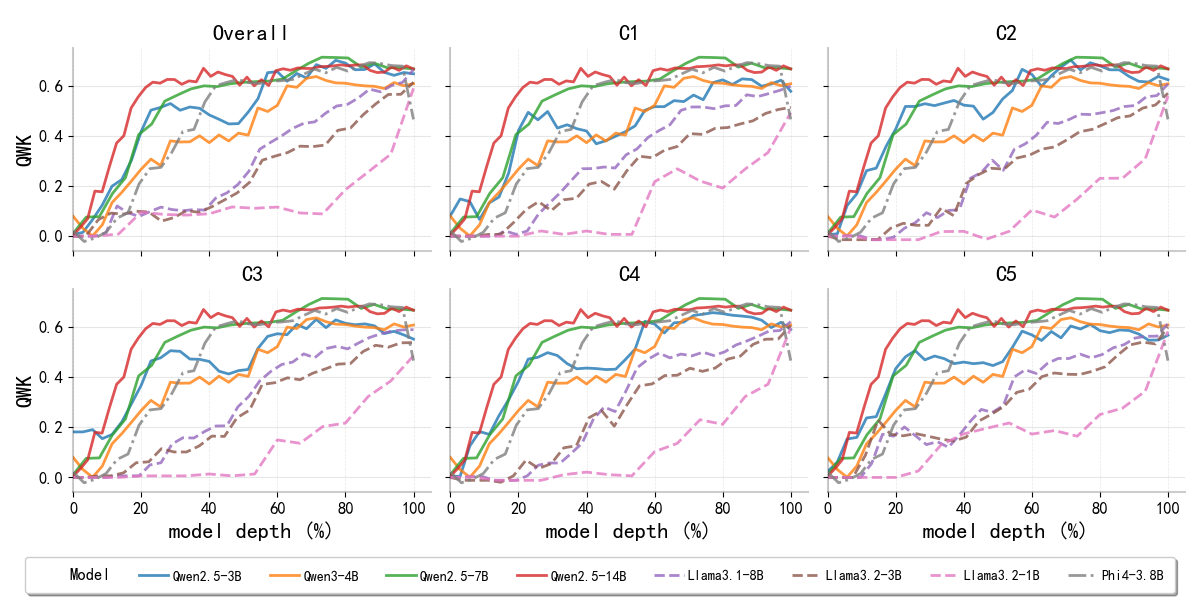}
    \caption{Average QWK scores of linear probes trained on ENEM across all essay prompts. Each subplot corresponds to a essay trait and shows probe performance across layers for different models.}
    \label{fig:ENEM}
\end{figure*}

\section{Results on CSEE and ENEM}
\label{sec:dataset_linear}
Figures~\ref{fig:CSEE} and~\ref{fig:ENEM} present the linear probe results on the CSEE and ENEM datasets, respectively. Both exhibit trends similar to those observed on the ASAP++ dataset (see Figure~\ref{fig:linear}).

Notably, as shown in Figure~\ref{fig:ENEM}, the probe curve exhibits greater fluctuations and a lower peak performance compared to the English-language datasets. This discrepancy may be attributed to the relatively smaller dataset size or the limited coverage of Portuguese-language training data.

\section{Spearman Correlation Results under Dimensionality Reduction}
\label{sec:pca_sp}
\begin{figure*}[h]
    \centering
\includegraphics[width=1\linewidth]{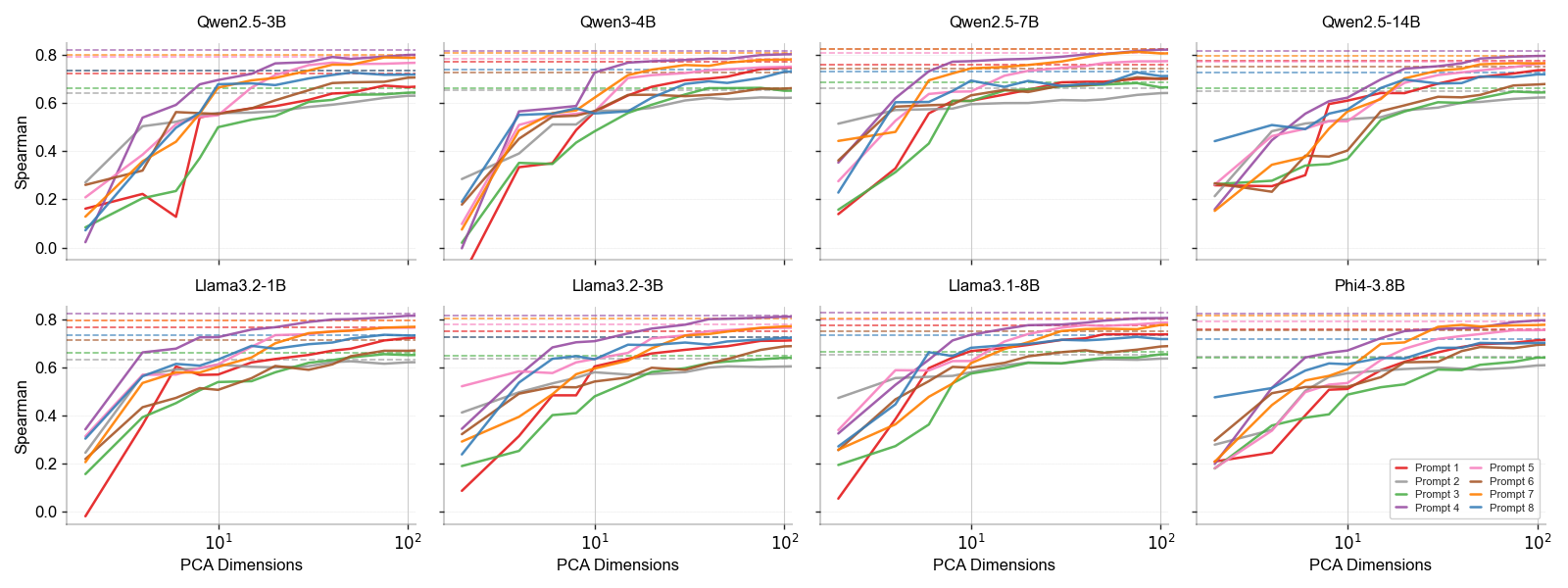}
\caption{Spearman correlation between predictions of probes trained on activations projected onto the top $k$ principal components and ground-truth scores. Each subplot shows results across different dimensionality reduction settings for each model, with the Spearman correlation of probes trained on full-dimensional activations shown as dotted lines.}

\label{fig:pca_sp}
\end{figure*}

Figure~\ref{fig:pca_sp} depicts the Spearman correlation between predictions of probes trained on activations projected onto the top $k$ principal components and ground-truth scores. Each subplot shows results across different dimensionality reduction settings for each model, with the Spearman correlation of probes trained on full-dimensional activations shown as dotted lines.

\section{Neuron Intervention}
\label{sec:Intervention}

To better understand the role of essay scoring neurons, we investigate the effect of intervening on a single essay scoring neuron ($L2.N392.W_{\text{in}}$, which exhibits a Spearman correlation of $0.574$ with Prompt 1 in ASAP++) in the Llama-3.1-8B-Instruct model.

Given a prompting template \( \mathcal{T} \) (see Figure~\ref{fig:llm_essay}), we fix the activation of this neuron across all tokens and sweep over a range of constant values, while tracking the prediction probabilities of the top-10 tokens (with \texttt{do\_sample=False}). As shown in Figure~\ref{fig:Intervention}, increasing the fixed activation leads to the largest increase in the weighted sum of essay-scoring neurons compared to three randomly selected neurons, while decreasing it produces the most pronounced decline. These results further suggest that this neuron is sensitive to intervention and plays a more prominent role in the model's essay scoring behavior than typical neurons.

\section{Neuron Distribution}
\label{sec:Neuron}
\begin{figure*}[h]
    \centering
    \includegraphics[width=1\linewidth]{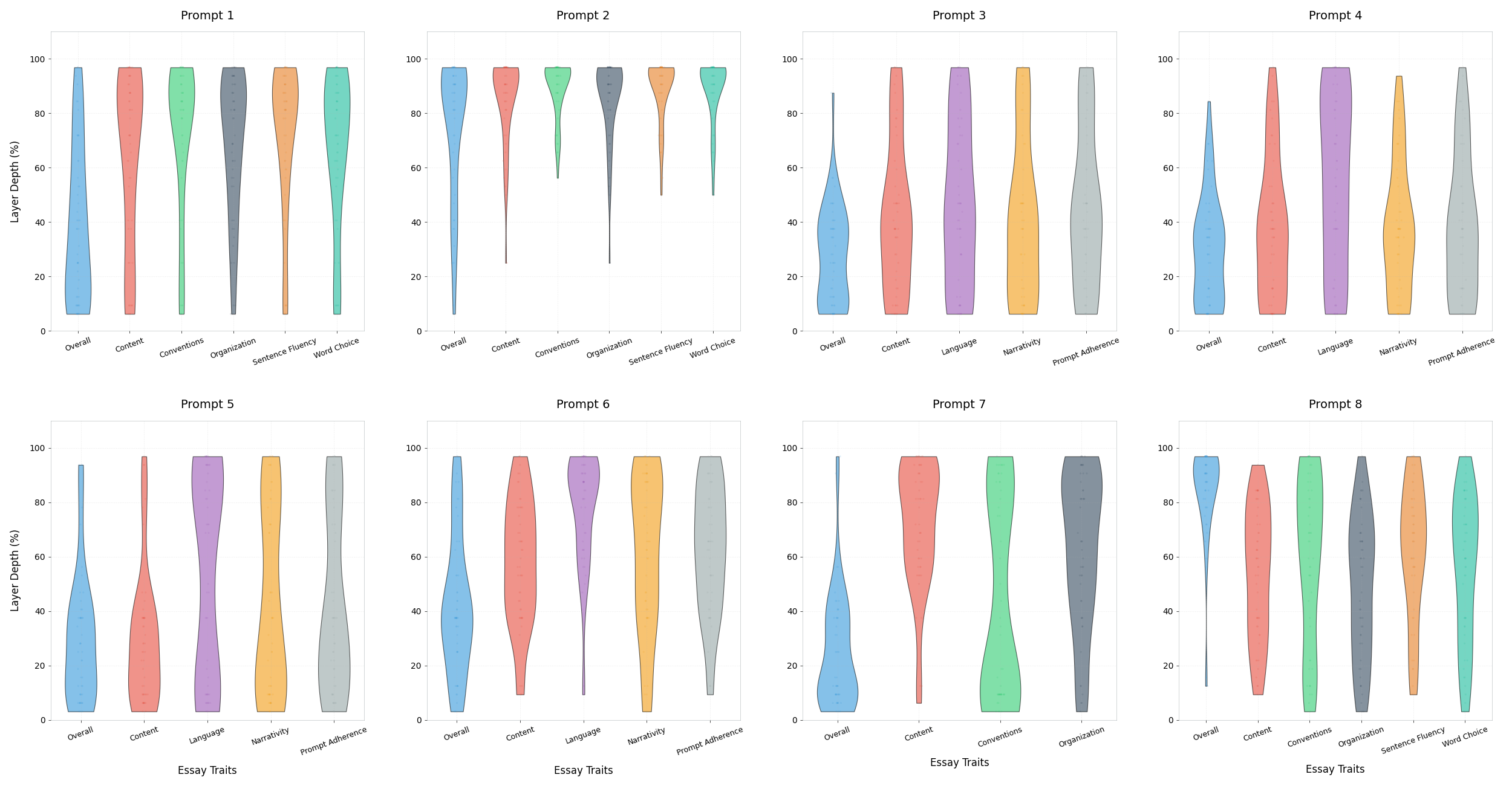}\hfill
    \includegraphics[width=1\linewidth]{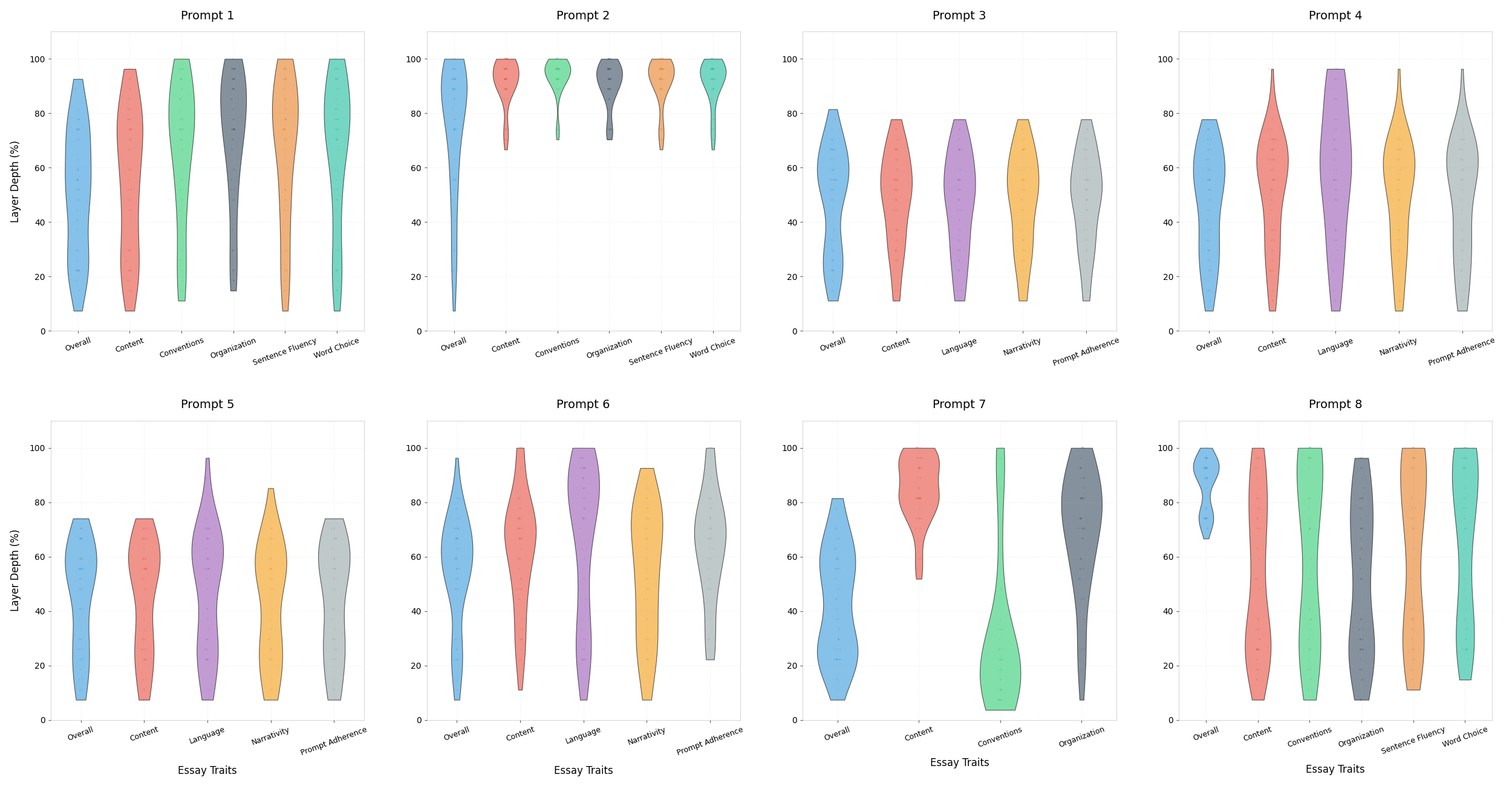}\hfill
    \includegraphics[width=1\linewidth]{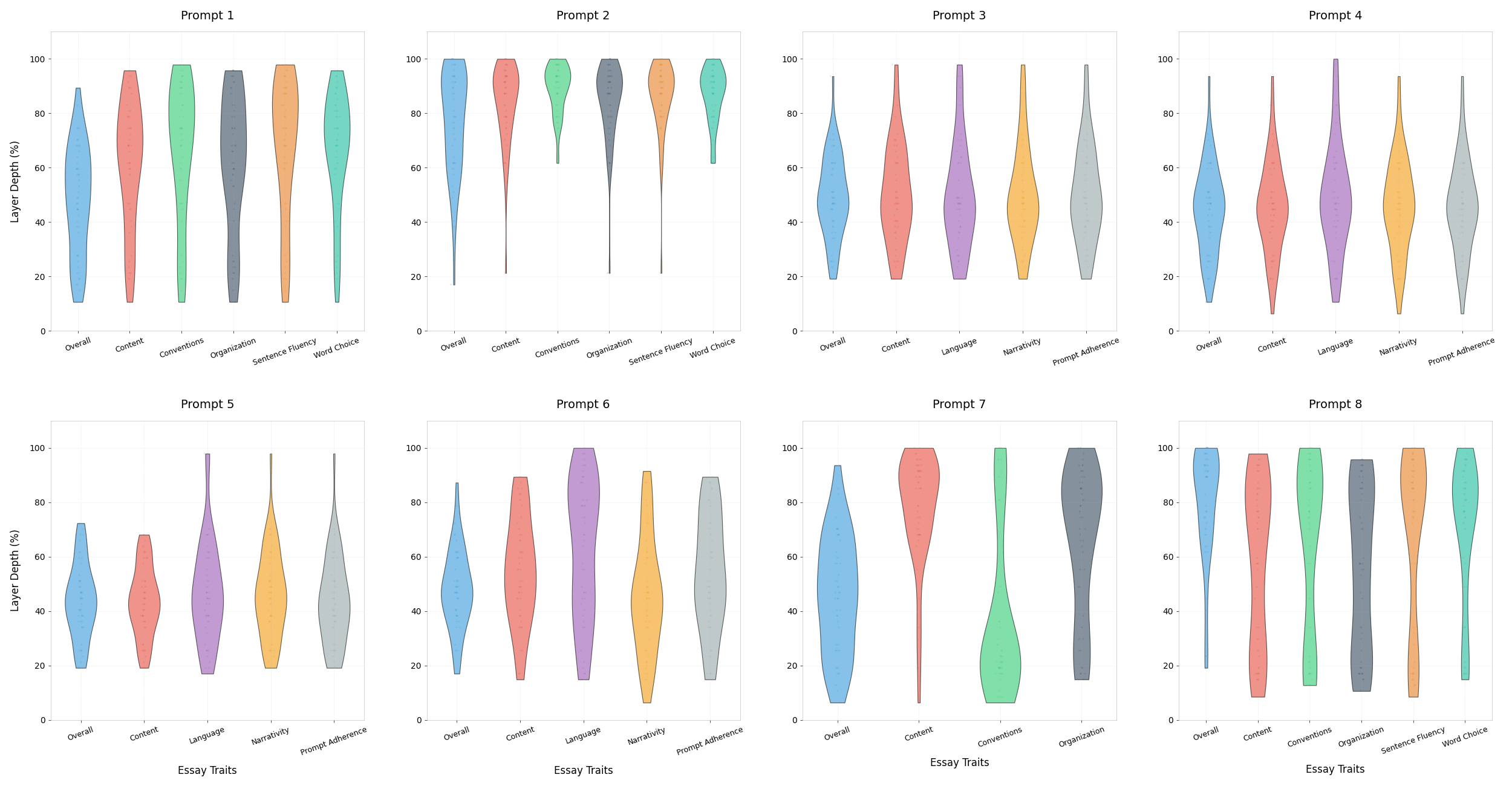}
    \caption{Distribution of the top 50 key neurons in the ASAP++ dataset, shown for different traits, essay prompts and models. Every two rows represent the results of a model, corresponding to Llama-3.1-8B-Instruct, Qwen2.5-7B-Instruct, and Qwen2.5-14B-Instruct, respectively.}
    \label{fig:layer_prompt}
\end{figure*}

Figure \ref{fig:layer_prompt} illustrates the distribution of the top 50 essay scoring neurons across different traits and essay prompts in the ASAP++ dataset, across models of varying architectures and sizes. All models demonstrate a consistent distribution across different essay prompts.

\section{Data Augmentation}
\label{sec:llm_essay}
To enable a statistically robust analysis of essay scoring neuron distributions across different essay lengths, we augmented Prompt 8 of the ASAP++ dataset with synthetic essays. The original dataset was insufficient to support reliable analysis at 100-word intervals, so we used the \texttt{gpt-5.4-mini} model to generate additional essays until each interval contained at least 100 samples. The temperature was set to 0.7 to encourage output diversity. The detailed generation prompt is provided below.

\begin{tcolorbox}[
    enhanced,
    colback=white,
    colframe=black,
    boxrule=0.8pt,
    arc=3pt,
    width=0.47\textwidth, % 稍微缩短整体宽度
    breakable,
]
\textbf{System Prompt}

\vspace{5pt}\hrule\vspace{5pt}

You are a helpful assistant and an expert in English language assessment. You will generate essays based on a given topic and score them according to the provided rubric.\\

\vspace{5pt}\hrule\vspace{5pt}

\textbf{User Prompt}

\vspace{5pt}\hrule\vspace{5pt}

**Essay Topic:** We all understand the benefits of laughter. For example, someone once said, "Laughter is the shortest distance between two people." Many other people believe that laughter is an important part of any relationship. Tell a true story in which laughter was one element or part.\\

**Your Task:**\\
1.  **Write the Essay:** Generate a short, true story based on the topic above. The story should be written from the perspective of a 10th-grade (Grade 10) student. The length must be between \{min\} and \{max\} words. (The average essay length for this topic is approximately 650 words.)\\
2.  **Score the Essay:** After writing, score the essay on a scale of 0 to 60 points. Use the following four criteria, each scored from 1 to 6 points, and note that Conventions has double weight:\\
- **Content (1-6 points):** This category assesses the core substance and clarity of a written piece. It focuses on how clear, focused, and well-supported the main ideas are.\\
- **Organization (1-6 points):** This category assesses the structure and flow of a piece of writing. It focuses on how logically and smoothly the ideas are ordered and connected for the reader.\\
- **Sentence Fluency (1-6 points):** This category assesses the rhythm, flow, and craftsmanship of sentences. It focuses on how smoothly and pleasantly the writing reads aloud, and the variety in sentence structure.\\
- **Conventions (1-6 points, double weight):** This category assesses the technical correctness of the writing, including grammar, punctuation, spelling, and capitalization. It focuses on how well the writer controls standard language rules to ensure clear communication.\\

**Scoring Formula:** Total = Content + Organization + Sentence Fluency + (2 × Conventions)\\

**Important:** The essay is short (\{min\}-\{max\} words), so scores should not be too high, and typically scores below \{max\_score\} points.\\

**Output Format Requirements:**\\
You must output **ONLY** a valid JSON object, nothing else. The JSON must have exactly two keys:\\
\{\{\\
    "essay": "Your generated essay text here..."\\
    "score": 12\\
\}\}
\end{tcolorbox}

\section{Base Model vs Instruction Tuned Model}
\begin{figure*}[b]
    \centering
\includegraphics[width=1\linewidth]{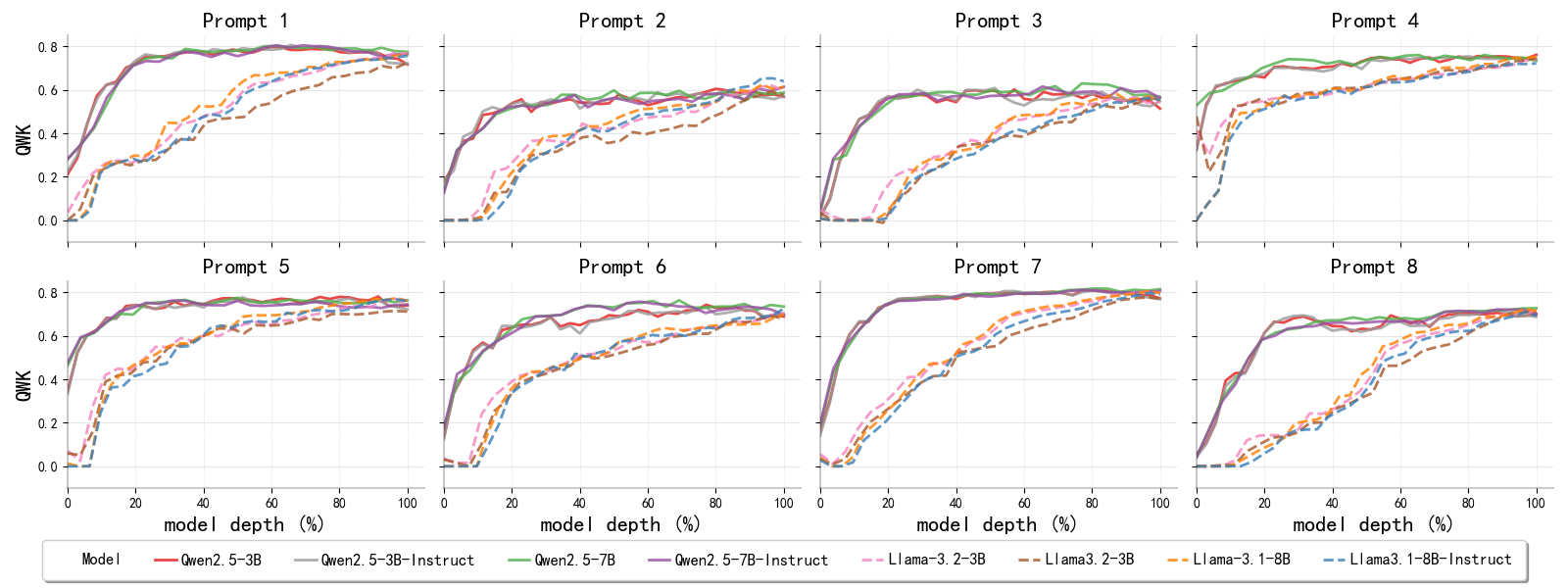}
    \caption{QWK scores of linear probes trained on the overall score of the ASAP++ dataset on each essay prompt and model.}
    \label{fig:base}
\end{figure*}

We investigate whether instruction tuning the base model enhances its capability to construct essay quality representations. As illustrated in Figure \ref{fig:base}, the performance of probes trained on the Base model and the Instruction tuned model exhibits negligible differences. This finding demonstrates that the model’s ability to construct representations of essay originates from the pretraining stage.

\begin{figure}[h]
\centering
\small % Reduce font size
\begin{tabular}{|p{0.45\textwidth}|} % Reduce column width
\hline
\textbf{Prompt $\mathcal{T}$} \\
\hline
Please score the following essay between 2 and 12 points, You only need to output the score.\\\relax
[Essay]\\
\{essay\}\\
(end of [Essay])\\
The Score is\\
\hline
\end{tabular}
\caption{Prompt template $\mathcal{T}$ used for neuron intervention experiments. Curly brackets \{\} denote placeholders to be completed.}
\label{fig:llm_essay}
\end{figure}

% \begin{figure}[h]
% \centering
% \small % Reduce font size
% \begin{tabular}{|p{0.45\textwidth}|} % Reduce column width
% \hline
% \textbf{Prompt $\mathcal{T}$} \\
% \hline
% Please score the following essay between 2 and 12 points, You only need to output the score.\\\relax
% [Essay]\\
% \{essay\}\\
% (end of [Essay])\\
% The Score is\\
% \hline
% \end{tabular}
% \caption{Prompt template $\mathcal{T}$ used for neuron intervention experiments. Curly brackets \{\} denote placeholders to be completed.}
% \label{fig:llm_prompt_template}
% \end{figure}

\end{document}